\documentclass[10pt,twocolumn,letterpaper]{article}

\usepackage{iccv}
\usepackage{times}
\usepackage{epsfig}
\usepackage{graphicx}
\usepackage{amsmath}
\usepackage{amssymb}
\usepackage{booktabs}
\usepackage[dvipsnames]{xcolor}
\usepackage[font=small,labelfont=bf]{caption}
\usepackage{icomma}
\usepackage[accsupp]{axessibility}

\usepackage[pagebackref=true,breaklinks=true,letterpaper=true,colorlinks,bookmarks=false]{hyperref}

\usepackage[capitalize]{cleveref}
\crefname{section}{Sec.}{Secs.}
\Crefname{section}{Section}{Sections}
\Crefname{table}{Table}{Tables}
\crefname{table}{Tab.}{Tabs.}

\iccvfinalcopy % *** Uncomment this line for the final submission

\def\iccvPaperID{11360} % *** Enter the ICCV Paper CD here

\begin{document}

%%%%%%%%% TITLE
%\title{AID: \underline{A}utomated \underline{I}maging System \underline{D}esign with Reinforcement Learning}
%\title{Joint Imaging System Design and Perception with Reinforcement Learning}
%\title{CoDIP: Co-Design of Imaging \& Perception with Reinforcement Learning}
%\title{DIPR: \underline{D}esigning \underline{I}maging Systems \& \underline{P}erception with \underline{R}einforcement Learning}
\title{DISeR: \underline{D}esigning \underline{I}maging \underline{S}yst\underline{e}ms with \underline{R}einforcement Learning}
%\title{CIPER: Co-design of Imaging \& Perception with Reinforcement Learning}
% \title{Joint Imaging System Design and Perception with Reinforcement Learning}
% Joint Perception and Imaging System Design with Reinforcement Learning
% 
% Automating Imaging System Design with Reinforcement Learning
% Co-design of Imaging Systems with Perception using Reinforcement Learning

\newcommand{\superscript}[1]{\ensuremath{^{\textrm{#1}}}}
\author{
    Tzofi Klinghoffer\superscript{*} \hspace{2mm} 
    Kushagra Tiwary\superscript{*} \hspace{2mm}   
    Nikhil Behari \hspace{2mm} 
    Bhavya Agrawalla \hspace{2mm}
    Ramesh Raskar\\
    %\vspace{2mm}
    Massachusetts Institute of Technology\\
    {\tt\small \{tzofi,ktiwary,behari,bhavya,raskar\}@mit.edu}
    %\normalsize{\textit{Project page}: \href{https://tzofi.github.io/diser/}{https://tzofi.github.io/diser}}
}

\maketitle
\def\thefootnote{* }\footnotetext{ Equal contribution.}\def\thefootnote{\arabic{footnote}}

%\ificcvfinal\thispagestyle{empty}\fi

% Bringing together the eyes and brains of perception
% Co-Designing the Eyes and Brains of Perception
% Autosensing: Automated Sensing for Perception
% Autosensing: Joint Capture and Perception with Reinforcement Learning
% Automating Imaging System Design with Reinforcement Learning

%%%%%%%%% ABSTRACT
\begin{abstract}
    Imaging systems consist of cameras to encode visual information about the world and perception models to interpret this encoding. Cameras contain (1) illumination sources, (2) optical elements, and (3) sensors, while perception models use (4) algorithms. Directly searching over all combinations of these four building blocks to design an imaging system is challenging due to the size of the search space. Moreover, cameras and perception models are often designed independently, leading to sub-optimal task performance. In this paper, we formulate these four building blocks of imaging systems as a context-free grammar (CFG), which can be automatically searched over with a learned camera designer to jointly optimize the imaging system with task-specific perception models. By transforming the CFG to a state-action space, we then show how the camera designer can be implemented with reinforcement learning to intelligently search over the combinatorial space of possible imaging system configurations. We demonstrate our approach on two tasks, depth estimation and camera rig design for autonomous vehicles, showing that our method yields rigs that outperform industry-wide standards. We believe that our proposed approach is an important step towards automating imaging system design. Our project page is \href{https://tzofi.github.io/diser/}{https://tzofi.github.io/diser}.

\end{abstract}

%%%%%%%%% BODY TEXT
\section{Introduction}

%In nature, animal sensing and processing evolve together to be optimal for survival in the animal's ecological niche. As a result, animal eyes and brains are tightly integrated.
%Furthermore, the perception systems of natural systems often make use of “unusual cues” present in their environment, leveraging as much of the environment as possible to accomplish necessary tasks for survival.

Cameras are ubiquitous across industries. In autonomous vehicles, camera rigs provide information on the ego-vehicle's surroundings so it can navigate; in biology, microscopy allows new viruses to be studied and vaccines to be developed; and in AR/VR systems, advanced headsets provide immersive reconstructions of the user's surroundings. In each of these applications, camera configurations must be carefully designed to capture relevant information for downstream tasks, often done with perception models (PMs). PMs are typically implemented as neural networks and use the output of cameras to predict information such as where other vehicles are on the road, what type of molecule is present in a biological sample, or where the user is located within a virtual environment. Yet, despite their interdependence, cameras and PMs are often designed independently. 

%imaging systems.

\begin{figure}
    \centering
    \includegraphics[width=\columnwidth]{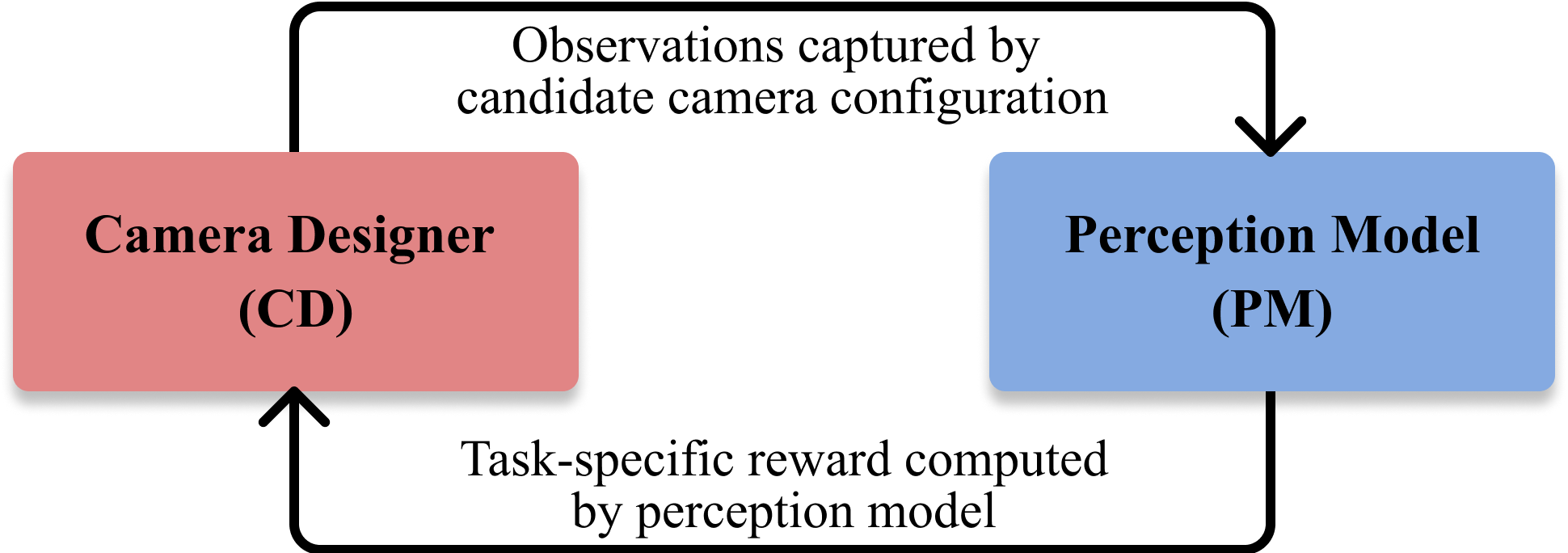}
    \caption{\textbf{Overview:} The camera designer selects imaging hardware candidates, which are used to capture observations in simulation. The perception model is then updated and computes the reward for the camera designer using the captured observations. In our paper, we implement the camera designer with reinforcement learning and the perception model with a neural network.}
    \label{fig:overview}
    \vspace{-4mm}
\end{figure}

\begin{figure*}
    \centering
    \includegraphics[scale=0.29]{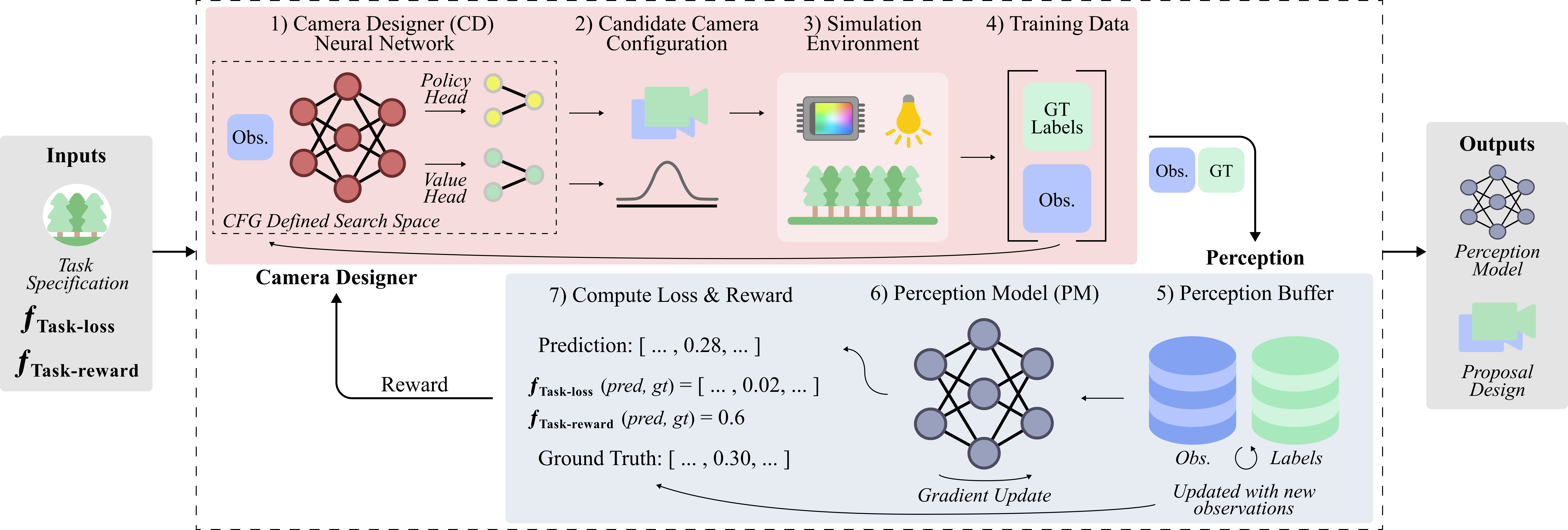}
    \caption{\textbf{Approach:} Our approach allows a camera configuration and perception model (PM) to be co-designed for task-specific imaging applications. At every step of the optimization, the camera designer (CD), implemented with reinforcement learning, proposes candidate camera configurations (1-2), which are used to capture observations and labels in a simulated environment (3-4). The observations and labels are added to the perception buffer (5) and used to compute the loss and reward, while the $N$ most recent observations in the perception buffer are used to train the PM. The reward is propagated to the CD agent which proposes additional changes to the candidate camera configuration. After the episode terminates, the CD agent is trained using proximal policy optimization (PPO) \cite{schulman2017proximal} until convergence.} %Optimization continues until task accuracy has been maximized.}
    \label{fig:architecture}
    \vspace{-4mm}
\end{figure*}

Designing camera systems is non-trivial due to the vast number of engineering decisions to be made. For example, consider designing a camera rig on an autonomous vehicle. Suppose the ego-vehicle is limited to up to $5$ lidar sensors, $5$ radars, and $5$ RGB sensors, with $1,000$ possible spatio-temporal resolutions. If there are $1,000$ discrete candidate camera positions on the ego-vehicle, the search space expands to $10^8$ different configurations. In practice, the search space can become many orders larger with more possibilities for each imaging system building block. Furthermore, because the search space is non-differentiable, there exists a need to develop efficient methods to effectively traverse the search space for an optimal imaging configuration.

In our paper, we propose using reinforcement learning (RL) to automate search over imaging systems. We first define a language for imaging system design using context-free grammar (CFG), which allows imaging systems to be represented as strings. The CFG serves as a search space for which search algorithms can then be used to automate imaging system design. We refer to such an algorithm as a camera designer (CD) and implement it with RL. RL allows us to search over imaging systems without relying on differentiable simulators and can scale to the combinatorially large search space of the CFG. Inspired by how animal eyes and brains are tightly integrated \cite{land2012animal}, our approach jointly trains the CD and PM, using the accuracy of the PM to inform how the CD is updated in training (Fig. \ref{fig:overview}). Because searching over the entire CFG is infeasible with available simulators, we take the first step of \textit{validating} that RL can be used to search over subsets of the CFG, including number of cameras, pose, field of view (FoV), and light intensity. First, we apply our method to depth estimation, demonstrating the viability of jointly learning imaging and perception. Next, we tackle the practical problem of designing a camera rig for AVs and show that our approach can create rigs that lead to higher perception accuracy than industry-standard rig designs. While AV camera rigs are one of many potential applications of our method, to the best of our knowledge, we are among the first to propose a way to optimize AV camera rigs. Our paper makes the following contributions:

\begin{itemize}
\item \textbf{Imaging CFG}: We introduce a context-free grammar (CFG) for imaging system design, which enumerates possible combinations of illumination, optics, sensors, and algorithms. The CFG can be used as a search space and theoretical framework for imaging system design.
\item \textbf{Co-Design}: We demonstrate how task-specific camera configurations can be co-designed with the perception model by transforming the CFG into a state-action space and using reinforcement learning (Fig. \ref{fig:architecture}). Our approach can converge despite the reward function being jointly trained with the policy and value functions.  
%reinforcement learning can be used to automatically search over the CFG to create task-specific imaging systems by reformulating the CFG as an action space.
\item \textbf{Experimental Validation}: We demonstrate our method for co-design by applying it to (1) the task of depth estimation using stereo cues, and (2) optimizing camera rigs for autonomous vehicle perception, showing in both cases that camera configuration and perception model can be learned together.

\vspace{-4mm}

%We show how our co-design framework can learn existing vision techniques, such as stereo depth estimation, and generate novel approaches to practical engineering problems, such as designing camera rigs for autonomous vehicles, that improve task performance.
%\item We show how our framework for automatic co-design of imaging systems and perception models can be used to solve vision problems, such as depth estimation, and solve practical engineering problems, such as designing camera rigs for autonomous vehicles.
\end{itemize}

\section{Related Work}

\subsection{Joint Optimization of Optics \& Algorithms}
Our work is most closely related to end-to-end optimization of cameras, which is an area of research focused on jointly optimizing components of cameras together with an algorithm, typically a neural network. Instead of relying on heuristics, the goal of end-to-end optimization is to produce images that optimize the pertinent information required for the task. Existing work primarily focuses on optimizing the parameters of the optical element, sensor, and image signal processor of a single camera. Applications of end-to-end optimization include extended depth of field and superresolution imaging \cite{sitzmann2018end}, high dynamic range (HDR) imaging \cite{metzler2020deep,sun2020learning}, demosaicking \cite{chakrabarti2016learning}, depth estimation \cite{baek2021polka,chang2019deep,haim2018depth,he2018learning}, classification \cite{chang2018hybrid} and object detection \cite{del2020learned,onzon2021neural,robidoux2021end}. Tseng \textit{et al.} \cite{tseng2021differentiable} employ gradient descent on a non-differentiable simulator by training a proxy neural network, whereas we directly operate on the non-differentiable simulator with RL. For a more comprehensive review of end-to-end optimization, we refer readers to \cite{klinghoffer2022physics}. In contrast to end-to-end optimization methods, we focus on optimizing over the much larger space of possible imaging system designs, rather than the parameters of an individual camera. Our search space contains varying illumination sources, optics, sensors, and algorithms, each with many parameters. Rather than using stochastic gradient descent for optimization, we use reinforcement learning, allowing our approach to be used with non-differentiable simulators.

%\subsection{Scientific Discovery}

%Scientific discovery is often driven by enumerating possible building blocks and interactions between those building blocks \cite{boyden2019architecting}. Using machine learning (ML) to accelerate the process of discovery is a natural tactic because it can allow faster searching over enumerations to find meaningful solutions. Scientific discovery has been accelerated by ML in many disciplines, such as protein folding \cite{jumper2021highly}, discovering methods for faster matrix multiplication \cite{fawzi2022discovering}, and AutoML \cite{bakerdesigning}. Often reinforcement learning (RL) is used as the ML machinery to engineer discovery. In the next section, we delve deeper into how RL can be used for scientific discovery. We end this section by noting that, like many of the other problems for which ML has helped accelerate discovery, CI contains a high dimensional space of possible combinations of hardware and software, many of which have not yet been explored. By defining an approach for using ML for designing CI systems, we hope our work can help future CI approaches by automatically discovered.

%strategy for using ML to search the space of possible combinations, we hope our contributions can help CI approaches to be automatically discovered.

\subsection{Reinforcement Learning}

Deep reinforcement learning (RL) has become widely used in recent years as a way to do sequential decision making for a wide array of problems, such as protein folding \cite{jumper2021highly}, learning faster matrix multiplication \cite{fawzi2022discovering}, and automated machine learning \cite{bakerdesigning}. Many RL techniques focus on the \emph{exploration-exploitation} trade-off, where an agent must learn to balance exploring new states with exploiting previously visited states that lead to high reward. RL is also used for many combinatorial optimization problems \cite{mazyavkina2021reinforcement}. In our work, we take inspiration from automated chip placement \cite{mirhoseini2021graph}, which, like our approach, is formulated to allow an RL agent to place a new component at every step and select the placement of that component. Like many other problems RL has been applied to, imaging contains a high dimensional search space. In our work, we use proximal policy optimization (PPO) \cite{schulman2017proximal}, which has been used for combinatorial search in past work~\cite{zhang2022deep}. 

Context-free grammars (CFGs) have been used to design machine learning (ML) pipelines, which are combinations of data-flows, ML operators, and optimizers~\cite{cfg1}\cite{cfg2}\cite{cfg3}. Typically, ML pipeline design is done via a search over strings in the CFG using tree search algorithms, such as Monte Carlo tree search or upper confidence trees \cite{uct} \cite{vazquez2022gramml}. CFGs have also been adopted for robot design \cite{zhao2022automatic}, molecule generation \cite{guo2022data}, and material design \cite{guo2022polygrammar}. We use CFG to functionally represent imaging systems as combinations of illumination, sensors, optics and algorithms such that the output string describes a camera configuration and perception model that can be used to solve a desired task. 

%Because many imaging simulators are not differentiable, gradient descent cannot be directly applied, but RL can be. 

%-----------------------------------------------------------------------
\section{Automated Imaging System Design}

%-------------------------------------------------------------------------
\subsection{Language for Imaging}
\label{sec:cfg}

% We propose a language to build imaging systems using context-free grammar (CFG). 

We define the configuration space of imaging systems using context-free grammar (CFG) as it allows for a flexible configuration space that can be searched. A typical context-free grammar, $G$, is represented as a tuple, $G=(V, \Sigma, P, R)$, where $V$ corresponds to non-terminal symbols in the grammar, $\Sigma$ corresponds to terminal symbols, $P$ corresponds to the production rules, and $R$ is the start symbol. The goal of our proposed CFG is to allow the construction of strings to represent arbitrarily complex imaging systems, which usually consist of illumination sources, optical elements, sensors to convert light into digital signals, and algorithms that decode the scene. For example, consider the task of depth estimation that can be done in numerous ways. One solution is depth from stereo, which involves placing two cameras, $c_1, c_2$, in the scene at points, $p_1, p_2$, with some baseline. Each camera has an optical element, $o_1 = (f,d)$, with a focal length, $f$, and aperture, $d$, and a sensor, $s_1 = ((h,w),t)$, with spatial and temporal resolutions, $(h,w)$ and $t$, respectively. Thus the cameras can be expressed as $c_1 = (o_1, s_1)$ and $c_2 = (o_2, s_2)$. An algorithm can decode the outputs of the two cameras to produce depth, and can be implemented with correspondence-matching~\cite{bradski2000opencv}, ($a_{st}$), or deep stereo~\cite{luo2016efficient}, ($a_{ds}$). The full system can be described as a string, $s_1 = ``c_1 c_2 a_{st}"$ or $s_2 = ``c_1 c_2 a_{ds}"$. Another way to estimate depth is with active illumination or time-of-flight (ToF) imaging. We can represent lidar as an algorithm, $a_{\text{control}}$, that illuminates the scene at the same point with a laser, $l_1$, and ToF sensor, $s_{ToF}$. We can describe this system as $s_{\text{lidar}} = a_{\text{control}} l_1 s_{ToF} a_{ToF}$. These examples illustrate how CFG can represent imaging systems with different illumination, optics, sensors, and algorithms as strings. The goal of the proposed CFG is not to describe how the individual components of an imaging system are made, e.g. their electronics, but rather to describe the function of each component. Next, we define the grammar's alphabet and production rules.

\begin{figure}
    \centering
    \vspace{-5mm}
    \begin{align}
        \mathrm{R} & \rightarrow \mathrm{X} \mathcal{S} \mathrm{X} \mathrm{A} \\ 
        \mathrm{X} & \rightarrow \mathcal{I} \mathrm{X} | \mathrm{O} \mathcal{S} \mathrm{X}| \mathrm{A}_2 \mathrm{X} | \epsilon \\
        \mathrm{O} &\rightarrow \mathcal{O} \mathrm{O} | \epsilon \\
        \mathrm{A}_1 &\rightarrow \mathcal{A}_1 \mathrm{A}_1 | \mathcal{A}_1 \\
        \mathrm{A}_2 &\rightarrow \mathcal{A}_2 \mathrm{O} \mathcal{S} | \mathcal{A}_2 \mathcal{I} | \mathcal{A}_2 \mathcal{S} | \epsilon \\
        % \noindent\rule{\textwidth}{1pt} \\
        \mathcal{S} &:= \{ s_{p} s_{hw} s_{t} s_{\lambda} s_{q} \}_{p \in \mathbb{R}^6,h,w,t,q \in \mathbb{Z} } \\
        \mathcal{O} &:= \{ o_{f} o_{d} \}_{f \in \mathbb{R} ,D \in \mathbb{Z}} \\
        \mathcal{I} &:= \{ i_{p} i_{i} \}_{p \in \mathbb{R}^6, i \in \mathbb{Z}} \\
        \mathcal{A}_1 &:= \{ a_{nn}, a_{fourier}, ... \} \\
        \mathcal{A}_2 &:= \{ \text{autofocus}, ... \} 
    \vspace{-5mm}
    \end{align}
    \vspace{-5mm}
    \caption{\textbf{Context-free grammar (CFG) for imaging:} Production rules (1-5) and alphabets (6-10) for our proposed CFG for designing imaging systems. $R$ is the starting symbol from which a design starts. All imaging systems must have at least one sensor, $\mathcal{S}$, and one algorithm, $\mathcal{A}$. The grammar allows arbitrary physically plausible combinations of illumination ($\mathcal{I}$) optics ($\mathcal{O}$), sensors ($\mathcal{S}$), and algorithms ($\mathcal{A}$), each defined in their respective alphabet above. $A_1$ refers to algorithms that process the output of hardware, while $A_2$ refers to algorithms that control hardware.}
    \label{fig:production_rules}
    \vspace{-4mm}
\end{figure}

\noindent
\textbf{Grammar.} Our proposed CFG can be stated as $G=(V, \Sigma, P, R)$. We define the variables as $V=\{\mathrm{X}, \mathrm{O}, \mathrm{A}_1, \mathrm{A}_2\}$, each defined in the following sections, and the terminals, $\Sigma$, which we refer to as alphabets, as $\Sigma=\{ \mathcal{I}, \mathcal{O}, \mathcal{S}, \mathcal{A}_1, \mathcal{A}_2 \}$, where $\{\mathcal{I}\}$ is illumination, $\{ \mathcal{O}\}$ is optics, $\{\mathcal{S}\}$ is sensors, and $\{\mathcal{A}_1\}$ and $\{\mathcal{A}_2\}$ are algorithms. Each alphabet contains possible components and parameters, defined in lower case, e.g. $a_{nn}$. Each component within an alphabet is parameterized by its functionality, e.g. focal length, rather than an off-the-shelf component. We describe each alphabet below and in Fig.~\ref{fig:production_rules}.

%\textbf{Grammar.} We define the following set of alphabets: Illumination $\{\mathcal{I}\}$, Optics $\{ \mathcal{O}\}$, Sensors $\{\mathcal{S}\}$, and Algorithms $\{\mathcal{A}\}$. Each set then contains alphabets pertaining to that class of the imaging module. For clarity, the alphabets, or terminals, are defined in script notation, $\Sigma = \{ \mathcal{I}, \mathcal{O}, \mathcal{S}, \mathcal{A} \}$, and the variables $V = \{\mathrm{X}, \mathrm{A}\}$. The individual alphabets are defined using a lower case, eg. $a_{fourier}$. Note that the alphabet set is a functional representation and is functionally parameterized i.e. its parameters describe its function not if the sensor exists in reality. The functional set can be reduced to only search over sensors that exist in reality. We now discuss the alphabet set. 

\noindent
\textbf{Illumination.} The illumination alphabet, $\mathcal{I}$, functionally represents different types of possible illuminations. In imaging, illumination can be represented with many parameters, such as duration $(d)$, intensity $(i)$, color, wavelength $(\lambda)$, polarization $\eta$, pose (position \& orientation), $(p)$ and modulation in space and time \cite{bhandari2022computational}. In the scope of this work, we consider pose and intensity. These can later be extended to other forms of illumination. 

\noindent
\textbf{Optics}. We define the optics alphabet, $\mathcal{O}$, to capture the most important (but not exhaustive) optical properties in an imaging system: focal length $(f)$ and aperture $(D)$. The optics alphabet can be extended to include more complex techniques such as phase masks or diffractive optical elements (DOE). The non-terminal $\mathrm{O}$ indicates that optical elements can be stacked to create a multi-lens system. 

%functionally represent the space of optics using parameters for focal length $(f)$, aperture $(D)$, and index of refraction $\eta$. 

\noindent
\textbf{Sensors.} The sensor alphabet, $\{\mathcal{S}\}$, functionally describes different types of sensors, such as RGB and SPAD. We parameterize a sensor by its pose $s_p$, spatial (or angular) resolution $s_{hw}$, temporal resolution $s_{t}$, bit quantization $s_q$ and wavelength $s_{\lambda}$. For example, a SPAD sensor has higher temporal resolution (picosecond scale) and generally lower spatial resolution (on the order of 1,000 to 100,000 pixels), while a typical RGB sensor (CMOS) has a higher spatial resolution (hundreds of megapixels), but a lower temporal resolution (30 fps). Similarly, quantization (for example) can be varied between 1, 8 or 12 bits. The pose is the position $(x,y,z)$ and the orientation (pitch, yaw, and roll) of the sensor in 3D space, $s_p \in \mathbb{R}^6$.

%Similar alphabets, Table 1 contains the alphabets defined for each of sets: Optics $\{\mathcal{O}\}$, Illumination $\{\mathcal{I}\}$, and Algorithms $\{\mathcal{A}\}$. 

\noindent
\textbf{Algorithms}. Algorithms are needed to decode raw images and control other alphabets. We denote the alphabet for algorithms with two sets: $\{ \mathcal{A}_1, \mathcal{A}_2 \}$. $\mathcal{A}_2$ is the set of algorithms that affect subsequent illumination, optics, and sensors (e.g. autofocus, controlling where to shine illumination), whereas $\mathcal{A}_1$ are algorithms that decode the incoming data from the sensors for a given task. These algorithms include standard imaging operators, such as the Fourier transform, backprojection, Radon transform, Gerchberg-Saxton algorithm, photometric stereo, and more. Additionally, $\mathcal{A}_1$ includes neural networks, which can perform detection, classification, etc. Due to the production rule, $\mathrm{A} \rightarrow \mathcal{A}_1 \mathrm{A} | \mathcal{A}_1$, $\mathcal{A}_1$ can be repeated and stacked together. For example, an algorithm can be designed that takes the Fourier transform of the input data and feeds it through a multilayer perceptron (MLP). 

\noindent
\textbf{Production Rules.} We define a set of production rules, shown in Fig.~\ref{fig:production_rules}, that can produce strings representing possible imaging system configurations. In our formulation, every imaging system includes at least one sensor and algorithm. The $\mathrm{X}$ accounts for imaging systems with different illumination, optics and sensors. In all cases, the string must end with at least one algorithm that outputs the desired task. Additionally, each $\mathcal{A}_2$ also requires an illumination, optics component, or sensor that it controls. The production rules account for multiple sensors and illuminations that illuminate and sense different parts of the scene. 

%For example, to estimate the depth of the point in the scene, a possible imaging configuration may use one illumination source, two sensors and an algorithm, $s_1 = ``i_{p_1}i_{10}(s_{p_2}s_{100,100}s_{30fps}s_{8bit})(s_{p_3}s_{100,100}s_{30fps}s_{8bit})a_{nn}"$. This configuration will force the algorithm to find correspondences from the outputs of the two sensors and triangulate those points to estimate depth. An alternative configuration may be $s_2 = ``A_2(i_{p_1}i_{10})s_{p_2}s_{100,100}s_{30fps}s_{8bit}a_{nn}"$. In this case, $A_2$ is used to control the lighting and hence estimate depth through structured light. 

%\KT{could also have a Task Set?}

% \input{tables/production.tex}

%-------------------------------------------------------------------------
\subsection{Imaging Design with Reinforcement Learning}

%The proposed context-free grammar (CFG) defines possible ways of combining illumination, optics, sensors, and algorithms to form an imaging system. We next propose the use of a learned imaging designer to automatically search over the CFG. In our work, we implement the imaging designer with reinforcement learning (RL). We use RL because it allows us to formulate optimization as a sequential decision making process, where at each step, the imaging designer can select which component from the CFG to add based on the task, environment, and current performance. We transform the CFG into an action space and model search as a Markov decision process (MDP). MDPs consist of four elements $<S, A, T, R>$:

The proposed context-free grammar (CFG) defines ways of combining illumination, optics, sensors, and algorithms to form an imaging system. The goal of our work is to automate imaging system design by searching over the CFG. Because the output of the cameras in the imaging system must be well suited for a specific, downstream task, we co-design them with the task-specific perception model (PM). We next propose using a learned camera designer (CD) to automatically search over the CFG. We implement the CD with reinforcement learning (RL) because (1) the combination of continuous variables in our CFG causes an explosion in the search space, which, as a result, makes search with methods such as Monte Carlo tree search (MCTS) \cite{browne2012survey} or alpha-beta search \cite{schaeffer1996new} intractable, and (2) many advanced imaging simulators are not differentiable \cite{geary2002introduction,unrealengine,haas2014history}, and thus gradient descent cannot be directly applied. Our problem is well suited for sequential decision making because the task performance achieved with each choice of camera configurations directly affects subsequent design choices.

\vspace{1mm}
\noindent
\textbf{Overview:} Our approach is illustrated in Fig.~\ref{fig:architecture}. The input is a task-specific loss and reward function. When optimization starts, the imaging system contains no hardware. At each step, the CD selects whether to add a component into the system and the component's parameters (Fig.~\ref{fig:architecture}a-b). A simulator can then be used to collect observations from the candidate camera configuration (Fig.~\ref{fig:architecture}c). These observations are used by the perception model to compute the reward and loss (Fig.~\ref{fig:architecture}4-7). The reward is used to train the CD and the loss is used to train the perception model. This loop repeats until a camera configuration and perception model have been created that maximize task accuracy.

\vspace{1mm}
\noindent
\textbf{RL Formulation:} We transform the CFG into a state-action space which the RL agent, henceforth referred to as the CD, can search over. We use proximal policy optimization (PPO) to train the CD and model the RL problem with the following states, actions, and rewards:

\begin{itemize}
\item states, $S$: the possible states of the world, which, in our case, are the possible enumerations of illumination, optics, and sensors, and possible observations that can be captured from each enumeration.
\item actions, $A$: the actions an agent can take at any step, which, our case, consist of choosing illumination, optics, sensors, algorithms, and all parameters.
% \item state transition, $T$: the process of moving from one state to another based on actions taken by the imaging designer, described as the probability distribution over next states given the current state and action.
\item reward, $R$: the reward for taking an action in a state, which, in our case, is computed by passing observations from the candidate camera configuration into the PM to compute accuracy for a target task.
\end{itemize}

\begin{figure}
    \centering
    \includegraphics[scale=1.2]{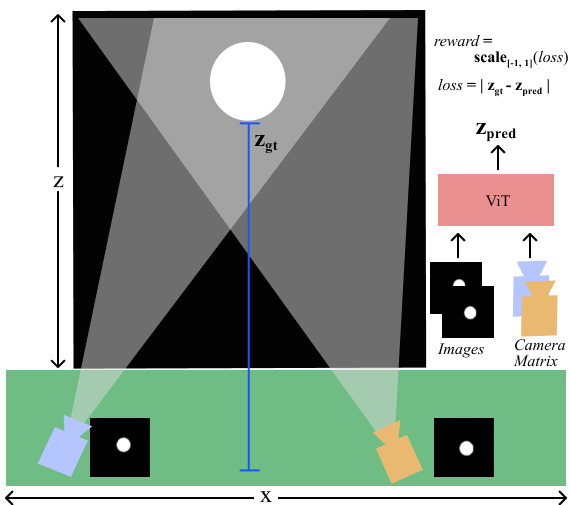}
    \caption{\textbf{Depth from Stereo Setup}: The goal of this experiment is to estimate the depth of a sphere using stereo cues. The camera designer (CD) places up to $C$ cameras within the green box. Camera poses and images are input to the perception model (PM) which outputs a predicted depth. We render environments that are devoid of monocular cues to force (1) the CD to learn to obtain multi-view cues and (2) the PM to learn to exploit these cues.
    % In theory, the CD could place a single camera and learn monocular cues (e.g. shading/lighting, texture, linear perspective), however, 
    % We test our formulation to jointly learn a camera configuration and a perception model to estimate depth using stereo cues. Our environment has no monocular cues to incentivize the learned system to predict depth using stereo or multi-view cues. The lack of monocular cues mean that the perception model will struggle to estimate depth from a single image, hence the agent must learn to place multiple cameras with a wide baseline to aid the perception model in its task of depth estimation. Moreover, the agent must bring the sphere into its FoV, and thus must place the cameras in a configuration that covers the space. The goal of this experiment is to understand if the camera designer and perception model can learn to estimate depth using two or more cameras. 
    }
    % \caption{\textbf{Depth Estimation Task:} We test our formulation to jointly learn an imaging setup and a perception algorithm for depth estimation using an environment with limited monocular cues. The task for the perception network is to estimate the depth to the sphere given observations from the imaging setup chosen by the agent. The depth estimation error serves as the reward for the agent to predict a better imaging setup in addition to the loss that updates the perception model weights. The policy and the perception networks are jointly trained from scratch to learn an imaging setup best suited to predict depth to the sphere. The environment is designed to ensure that relying only on monocular cues will result in in-accurate depth estimation.}
    \label{fig:tri_cartoon}
    \vspace{-4mm}
\end{figure}

\noindent
\textbf{Simulation \& Environment}: Unlike standard RL problems where the agent acts based on observations from a fixed sensor, the observations provided to the CD can change, meaning the CD has to learn how to act with varying input (e.g. varying numbers of images, sensor parameters, etc). The simulator should thus be able to render data from all potential imaging systems that can be derived from the CFG. Because simulators that encompass the entire CFG are not available, we search over subsets of the CFG to validate our method. While we use a simulator, a dataset can also be used with offline RL approaches~\cite{levine2020offline}.

%using simulators such as PyRedner \cite{pyredner} and CARLA \cite{dosovitskiy2017carla}

%Such diverse imaging configurations are typically not available in standard RL simulators such as MuJoCo \cite{mujoco}. To enable a wider set of observation and sensor modality spaces, we employ PyRedner \cite{pyredner} and CARLA \cite{dosovitskiy2017carla}. 

%containing observations from different sensors. Such a dataset can be
%\emph{how} to observe, in addition to \textit{what} to observe and \textit{how} to act. For example, our agent can choose to observe the scene with two or three cameras with varying camera configurations (ToF, different field-of-view) which will change \textit{what} it observes in the scene. The simulator should also simulate second and third-order imaging effects such as shadows, or reflections and sensing modalities such as polarization. Such diverse imaging configurations are typically not available in typical RL simulators such as Mujoco \cite{mujoco}. To enable a wider set of observation and sensor modality spaces, we employ PyRedner \cite{pyredner} and Carla \cite{dosovitskiy2017carla}. While we use a simulator, a dataset can also be collected  containing observations with different sensors in an environment. Such a dataset can be used together with offline RL approaches \cite{levine2020offline}.

\vspace{1mm}
\noindent
\textbf{Perception Model:} In our experiments, we set the algorithm, $\mathcal{A}$, to be a trainable neural network (NN). The NN's role is to produce a task prediction given arbitrary observations from candidate camera configurations. The NN must be able to map a varying input (number of observations, modality, etc.) to a fixed output. For example, the CD may increase the number of sensors in the system beyond one, leading to multiple observations. We propose using transformers to mitigate this problem since they map a dynamic number of observations to a fixed-size feature embedding by converting inputs into sequences of patches~\cite{selva2022video}. To reduce noise in the gradients when jointly training the PM with the CD, we propose a perception buffer (Fig.~\ref{fig:architecture}.5), which stores the previous $N$ observations from candidate camera configurations, allowing the PM to be trained over all data in the buffer at each step. 
\section{Experiments and Results}
\label{sec:experiments}

\noindent
\textbf{Overview:} We apply our method to two problems, both of which exercise a subset of our proposed CFG to validate DISeR. First, we show how DISeR can jointly learn a camera configuration and perception model to solve depth estimation. Second, we apply DISeR to a practical engineering problem of designing camera rigs for AVs. The same formulation is used in both problems: at each step of optimization, the CD chooses whether to add a camera to the imaging system by predicting an action, $p$, in $[0, 1]$, referred to as camera placement probability, along with camera parameters. When $p$ is greater than a threshold of $0.5$, a camera is added with the predicted parameters. The camera parameters for each problem are shared in the sections below. In both problems, we compare our approach against random search, which we note is often very difficult to beat \cite{bergstra2012random} \cite{zoph2016neural}.

\subsection{Stereo Depth Estimation}
\label{sec:stereo}

\begin{figure}
    \centering
    \includegraphics[scale=0.23]{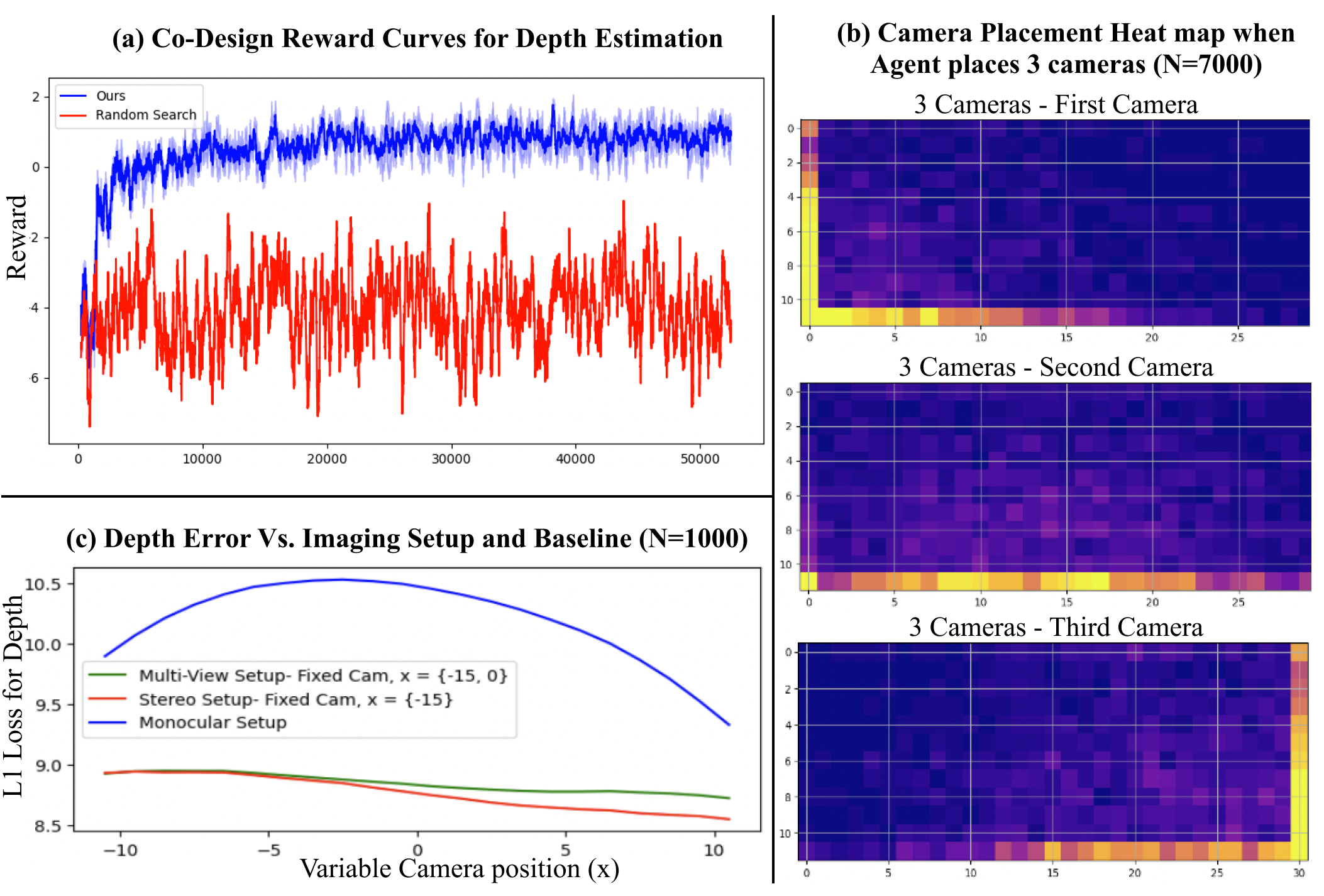}
    \caption{\textbf{Joint Camera and Perception Design for Stereo Depth.} We train the CD and PM from scratch to estimate depth of a sphere. (a) Our reward function consistently improves, even though it constantly changes due to the PM concurrently training with the CD. (b) The CD learns to maximize the baseline between different cameras over the course of $1000$ experiments when placing $3$ cameras. (c) The loss decreases with more placed cameras and larger distances between the cameras, which shows that the PM learns to exploit multi-view cues.}
    \label{fig:tri_eval}
    \vspace{-4mm}
\end{figure}
% \vspace{-3mm}

\subsubsection{Experimental Setup}

\paragraph{Environment} The goal of the first experiment is to estimate the depth of a sphere using stereo cues. The CD is allowed to place a maximum of $C$ cameras in the scene (though it can also place fewer cameras). In theory, the CD could place a single camera and learn monocular cues (e.g. shading/lighting, texture, linear perspective). However, we simulate an environment where monocular cues are unavailable, making monocular depth estimation ill-posed. 

Our environment consists of a randomly placed white sphere with a random radius, as shown in Fig.~\ref{fig:tri_cartoon}. We use PyRedner \cite{pyredner} to render images. The sphere position and radius are randomly sampled per episode from $(r, x, z) = \{r\in [3,9], x\in[-10, 10], z\in [1, 60] \}$. The depth is the $z$ distance from the sphere to the average position of the placed cameras. The scene is illuminated such that shading cues and the position of the light source are absent as cues. The only feedback that the PM and CD receive is a loss between the predicted and ground truth depth. The goal of rendering such an environment is to determine whether the CD can adapt to the context and realize that only a multi-view system can estimate depth. In parallel, the PM learns to exploit multi-view stereo cues. We show the supervised results of this experiment for validation in the supplement. 

% predict depth using stereo or multi-view cues. In our setup, an agent can place multiple cameras to observe the scene while a neural network is jointly trained to predict depth given the observations (which may or may not contain the sphere).  The lack of monocular cues mean that the perception model will struggle to estimate depth from a single image, so the imaging designer must learn to place multiple cameras with a wide baseline to aid the perception model in depth estimation. Moreover, the agent must bring the sphere into its FoV, and thus must place the cameras in a configuration that covers the space. The goal of this experiment is to understand if the imaging designer and perception model can learn to estimate depth using two or more cameras.  

% Our goal of this experiment is to jointly learn an imaging setup by jointly learning a hardware configuration and a perception model that estimates depth to the sphere. This experiment is an example of how our framework to jointly learn hardware and algorithm configurations can lead to learning existing hardware and vision configurations such as a setup that uses multiple cameras to estimate depth from stereo. 

% Our environment is designed in a way such that in order to predict accurate depth, the imaging designer must recognize that the task is to estimate the depth to the sphere and learn to place cameras in a configuration that enables the perception model to estimate accurate depth. 

\vspace{1mm}
\noindent
\textbf{Action Space:} The action space for depth estimation is  $(p, x, z, \theta) = \{p \in [0,1], x \in [-15, 15], z \in [69, 80], \theta \in [-60^{\circ}, 60^{\circ}] \}$, where $p$ is camera placement probability, $(x, z)$ is location (see Fig.~\ref{fig:tri_cartoon}) and $\theta$ is yaw. FoV is $45^{\circ}$. %set to 

%Our search space is a subset of the CFG described in Section~\ref{sec:cfg}. It includes the sensor positions and parameters of the neural network. At each step, the CD chooses to place a camera with probability $p$, at location $(x, z)$ with yaw angle $\theta$. The action space consists of the configuration of a camera: $(p, x, z, \theta) = \{p \in [0,1], x \in [-15, 15], z \in [69, 80], \theta \in [-60, 60] \}$. If $p>0.5$, the CD places the camera with the selected parameters and a fixed FoV of $45^{\circ}$. For each camera $i$, the camera matrix $\mathbf{P}_i$ is calculated and an observation $\mathbf{o}_i$ is captured. The final state is $ S = \{(\mathbf{o}_i, \mathbf{P}_i) \}_{i=1}^N$ is input to the perception model and CD, where $N$ is the number of cameras placed by the CD.

\vspace{1mm}
\noindent
\textbf{Experiment Details}: We use a modified version of the vision transformer (ViT) architecture \cite{dosovitskiy2020image} \cite{Arnab2021ViViTAV} that accepts an arbitrary number of images of fixed resolution and their corresponding camera parameters as input, and outputs a scalar depth. The spatial resolution is fixed to $(128, 128)$. The maximum number of cameras the CD can place is set to $C=5$. The CD's PPO backbone and the perception model share the same network architecture and are initialized randomly. The reward is computed before updating the perception model and is re-scaled to $[-1, 1]$. Additional information about the training is provided in the supplement. 
% The perception buffer size is set to 35 images and the perception network is updated twice every step. PPO is parallelized over 5 processes, and the agent is updated every 256 steps. 
\begin{figure*}
    \centering
    \includegraphics[scale=0.3]{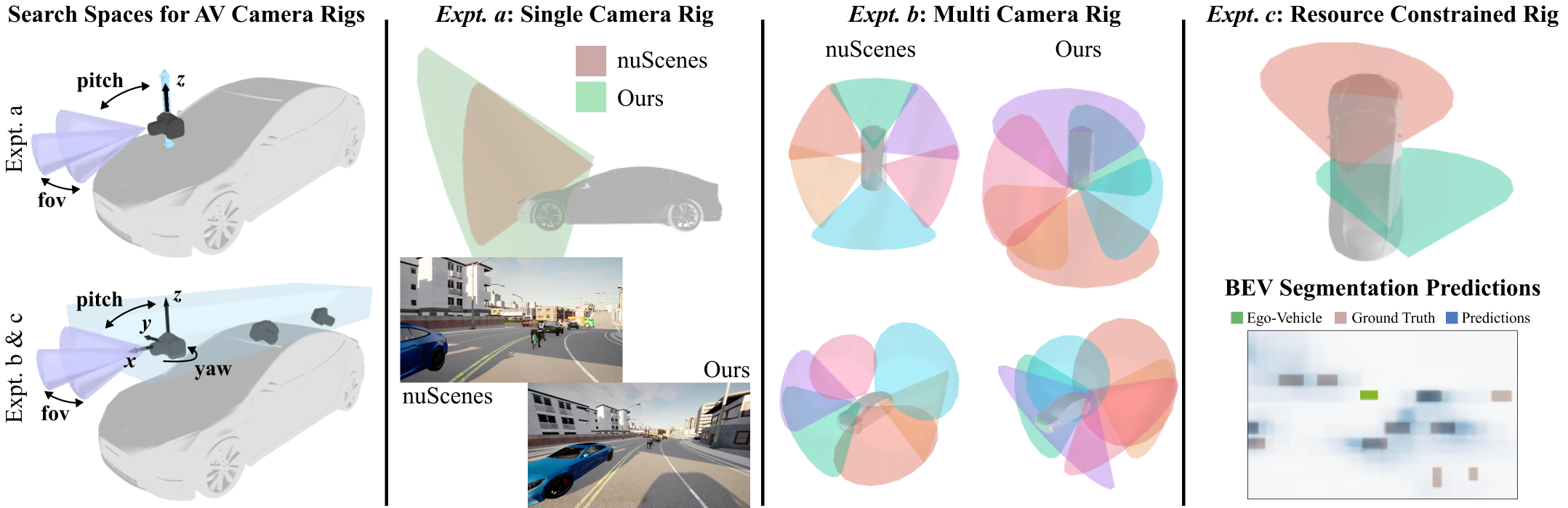}
    \caption{\textbf{Autonomous Vehicle (AV) Camera Rig Task \& Results:} We demonstrate that our approach can be used to create AV camera rigs that are optimized for BEV segmentation. (Left) Our search space is shown -- in expt. a, we optimize the height, pitch, and FoV of a single camera rig, while in expt. b and c, we optimize \# cameras, x, y, z, pitch, yaw, and FoV. Results for each experiment are shown and we compare the optimized camera rig to the camera rig used in nuScenes \cite{caesar2020nuscenes}. In expt. c, the camera designer learns to place fewer cameras in only the direction where cars are placed. We also show the BEV segmentation predictions of our jointly trained perception model.}
    \label{fig:carla_overview}
\end{figure*}
% Our environment is designed to force the CD and perception model to co-learn to find ways to estimate depth when limited monocular cues are available. 
%the search space is illustrated -- the CD selects how many cameras and what camera intrinsics and extrinsics are used for either a single camera or multi-camera rig. (Middle) We compare the rigs created with our approach with standard rigs, such as the one from the nuScenes dataset, and find that (Right) our approach results in higher BEV segmentation confidence predictions.

\subsubsection{Results and Discussion}

We evaluate the joint training (Fig.~\ref{fig:tri_eval}a), the learned policy (Fig.~\ref{fig:tri_eval}b), and the perception model (Fig.~\ref{fig:tri_eval}c) in isolation. Fig.~\ref{fig:tri_eval}a illustrates how our system maximizes reward when co-designing the PM with the camera design. The reward function is dictated by the output of the PM, but the PM is concurrently training with the camera design, which results in inconsistent rewards during training for the same states. In spite of this fact, our model is able to consistently increase the reward, even at the beginning of training when the PM is untrained and with random initialization. Our results show that the CD and PM are able to learn intuitions that hold true in conventional multi-view stereo.

\paragraph{Strategy \#1 -- Maximize Coverage:} When given the option to place up to $5$ cameras, the CD places $1$ camera $7.6\%$ of the time and 2, 3, 4, and 5 cameras $27.7\%$, $36.6\%$, $22.7\%$, $5.4\%$ times, respectively. Fig.~\ref{fig:tri_eval}b shows the heatmaps of where the CD decides to place each camera, specifically when the CD chose to place exactly three cameras. The heatmaps denote the number of times the CD placed the camera at a particular location over the course of $7000$ experiments, where each experiment denotes the placement of a new random size sphere at a random location. From the heatmaps, we see that the CD strategically placed the cameras at locations that maximize the baselines between different cameras. Camera 1 was predominantly placed in the left side of the allowed region, camera 2 at the center bottom, and camera 3 at the right. From these results, we see that the CD optimizes to place more cameras spaced far apart. However, placing more cameras doesn't necessarily mean that the CD is obtaining multiple views of the object (e.g. some cameras may be pointed in the opposite direction of the object). Therefore, we account for this case by defining the metric of \emph{coverage}, which defines the number of cameras that have at least one pixel viewing the object. The CD policy learns a configuration which maximizes coverage of the allowed region. We find that performance improves as coverage increases from $0$ to $3$, with the L1 loss being 14.0, 9.2, 7.2, and 5.7 as the coverage increases. Coverage is discussed in detail in the supplementary.  

% \vspace{-2mm}
\vspace{1mm}
\noindent
\textbf{Strategy \#2 -- Multi-View Cues and Maximal Baseline:}  Fig.~\ref{fig:tri_eval}c shows that the PM learns to exploit stereo cues when presented with multiple images. The experiment shown here compares the PM performance on a one-camera, two-camera, and three-camera system when estimating the depth of a sphere (averaged over $1000$ different spheres of varying size and depth). All three systems have a camera that can be moved along the $x$ axis, the two- and three-camera system have a fixed camera at $x=-15$, and the three-camera system has an additional fixed camera at $x=0$. The blue curve illustrates the L1 loss between the ground truth and one-camera system predictions. The red and green curves illustrate the performance of the two-camera and three-camera system respectively. The three-camera system performs slightly better than the two-camera system, and both perform significantly better than the one-camera system. The multi-view systems also see a decrease in loss (and variance) as the baseline between the cameras increases (i.e. as the movable camera moves along the $+x$ axis). These curves indicate that the PM has learned similar wisdom to that of conventional stereo -- multiple views with a large baseline enable better depth estimation~\cite{bhandari2022computational}. While gradient descent could also be used to learn to maximize baseline given a differentiable simulator, we use RL, which can be used with non-differentiable simulators to search over both number of cameras and their baseline.

\vspace{1mm}
\noindent
\textbf{Searching Illumination:} We also repeated the above experiment with an expanded action space that includes angle and intensity of a single spot light at a fixed position. To estimate the depth of the sphere, the CD must learn to sweep the light over the scene with a sufficiently high intensity until the sphere is illuminated. At each step, angle and intensity can be changed within the bounds of $[-60^{\circ}, 60^{\circ}]$ and $[0, 1]$, respectively, where 0 leaves the scene dark and 1 illuminates it. We found that the CD learns to increase the intensity so the sphere can be illuminated and change the angle such that the number of illuminated pixels on the sphere consistently increases over the episode.

\subsection{Camera Rigs for Autonomous Vehicles}

\begin{figure}
    \centering
    \includegraphics[width=\columnwidth]{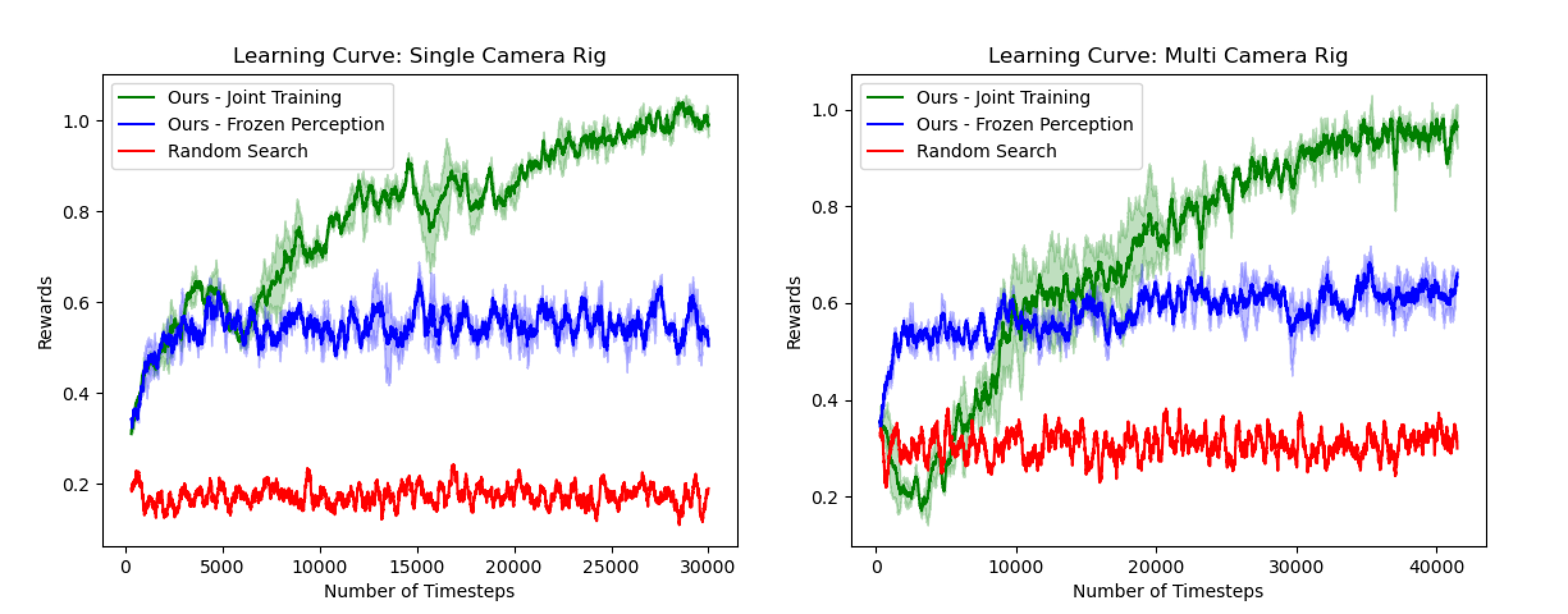}
    \caption{\textbf{Results for AV Camera Rig Co-Design:} Shown are the reward curves for the CD optimizing camera rigs for BEV segmentation. Reward is intersection over union (IoU). To demonstrate the effectiveness of co-designing the camera configuration with the perception model (PM), we show results when the PM is pre-trained and frozen (blue) vs. pre-trained and fine-tuned (green). Compared to random search (red), where actions are uniformly sampled from a random distribution at each step, our approach significantly outperforms, and discovers camera rigs that increase BEV segmentation IoU.}
    
    %We show the imaging designer's training rewards, measured as intersection over union (IoU) for BEV segmentation, for two variations of the task: designing a single camera rig (left) and designing a multi-camera rig (right). We compare against random search and show that the imaging designer, implemented with reinforcement learning, significantly outperforms and is able to learn to create camera rigs that increase the perception model's accuracy.}
    \label{fig:carla_results}
    \vspace{-4mm}
\end{figure}

%Our goal is to find a camera rig that leads to optimal performance for AV perception tasks. 

Next, we describe how our method can be used to optimize an AV camera rig for the perception task of bird's eye view (BEV) segmentation by jointly training the CD and PM. We validate our approach with three sets of experiments, described in the Experiment Details section below. We find that the rigs created with our approach lead to higher BEV segmentation accuracy in our environment compared to the industry-standard nuScenes \cite{caesar2020nuscenes} rig. Our camera rig search space and results are visualized in Fig. \ref{fig:carla_overview}.

%We describe how our method can be used to solve practical engineering problems that rely on complex imaging systems and perception models, such as optimizing an autonomous vehicle (AV)'s camera rig for AV perception tasks. We use the CD to search over a subset of the proposed CFG which includes the number of cameras and camera parameters such as position, orientation, and field-of-view. As a motivating task, we optimize the camera rig and perception model jointly for the task of bird's eye view (BEV) segmentation and show that our method can generate camera rigs that lead to higher performance than industry-wide standards, such as the nuScenes \cite{caesar2020nuscenes} rig. In addition, we show hand-crafted test scenarios and penalties can be used to train the CD to generate rigs that enable perception in specific conditions, such as different distributions of traffic or resource constrained camera rigs.

\subsubsection{Experimental Setup}

\vspace{1mm}
\noindent
\textbf{Environment:} We use the CARLA Simulator~\cite{dosovitskiy2017carla} to render observations from candidate camera rigs selected by the CD during training. For every camera on the candidate rig, the environment returns images, extrinsics, intrinsics, and 3D bounding box labels of vehicles in the scene. The 3D bounding boxes are used to compute the reward (for training the CD) and loss (for fine-tuning the PM). We use the same CARLA environment to create 25,000 samples rendered from randomly generated camera rigs to pre-train the PM for the task of BEV segmentation.

%A dataset of 25,000 samples from random camera rigs is rendered and used to first pre-train the BEV segmentation network. Each camera rig used for pre-training contains six cameras with random FoV, pitch, yaw, $x$, $y$, and $z$ in the intervals defined above. 

%The loss is used to fine-tune the perception model and the IoU is used as reward. The imaging designer then uses the reward and observations to update its policy.

\vspace{1mm}
\noindent
\textbf{Action Space:} The action space for AV camera rig design is $(p,x,y,z,\theta,\beta,\lambda)=\{p \in [0,1], x \in \eta_x, y \in \eta_y, z \in [z_{max}, z_{max} + 0.5 m], \theta \in [-180^{\circ}, 180^{\circ}), \beta \in [-20^{\circ}, 20^{\circ}], \lambda \in [50^{\circ}, 120^{\circ}] \}$, where $p$ is the camera placement probability, $(x,y,z)$ is location, $\theta$ is yaw, $\beta$ is pitch, and $\lambda$ is FoV. $\eta_x$ and $\eta_y$ are the extents of the ego-vehicle in x and y, respectively, and $z_{max}$ is the height of the ego-vehicle, meaning cameras can be placed anywhere within 0.5 meters (m) above the ego-vehicle. This action space conforms with rooftop rigs used in industry and the size and height of the roof match the Renault Zoe from nuScenes.

\begin{table}[t]
\centering
\begin{tabular}{p{0.27\linewidth}p{0.27\linewidth}p{0.27\linewidth}}
\hline
 & IoU (Expt. a) & IoU (Expt. b) \\
% & @0.40/@0.50 & @0.40/@0.50 \\
\hline
Random Rig & 0.254 & 0.084 \\
nuScenes Rig & 0.267 & 0.355 \\
Our Rig & \textbf{0.341} & \textbf{0.427} \\
%nuScenes Rig & 0.324 / 0.267 & 0.429 / 0.355 \\
%Our Rig & \textbf{0.377 / 0.341} & \textbf{0.452 / 0.427} \\
\hline
\end{tabular}
\captionof{table}{We compare the BEV segmentation IoU for models trained and tested with a random rig, nuScenes rig, and our approach's rig. CARLA train and test scenes are the same for each. Our rig achieves higher performance than industry standards.}
\label{table:iou}
\vspace{-4mm}
\end{table}

\vspace{1mm}
\noindent
\textbf{Experiment Details:} 
We use a recent BEV segmentation model, Cross View Transformers (CVT) \cite{zhou2022cross}, as the PM. It is first pre-trained on a dataset containing randomly placed cameras to allow it to more easily generalize to all candidate camera rigs that the CD may select. We then use the pre-trained CVT model to initialize the PM and CD's PPO backbone. Finally, we train the CD to optimize camera rigs. The PM uses the observations from each candidate rig and 3D bounding box labels to compute a reward (IoU) and loss (binary segmentation loss). The reward is used to update the candidate camera configuration, while the loss is used to update the PM. We conduct three sets of experiments, one with a single camera rig (expt. a), one with a multi-camera rig (expt. b), and one with a custom scenario and penalty for placing many cameras (expt. c). Each experiment is conducted with a frozen and a jointly trained PM. Fig.~\ref{fig:carla_results} shows that joint training leads to higher rewards (IoU).

%can be used to update the perception model, though we show that this is not necessary in our experiments. 
%We train the CD with a frozen perception model and a jointly trained perception model and show the benefit of joint training in Fig. \ref{fig:carla_results}, with higher IoU being achieved when with joint training. 

%\noindent
%\textbf{ -- Expt. a:} The imaging designer selects where to place a single camera in the front part of the ego-vehicle, selecting only the pitch, height, and FoV. We use the same formulation as described above, but, at each step, if the imaging designer places a camera, the previous camera is \emph{replaced}, rather than the new one being added on the rig. After six steps, the episode terminates.

%\noindent
%\textbf{ -- Expt. b:} The imaging designer selects both how many cameras and the parameters to instantiate them with. The imaging agent is restricted to place cameras anywhere on top of the ego-vehicle within its extents. Again, we set the episode to terminate after six steps, meaning a maximum of six cameras can be placed.

%\noindent
%\textbf{ -- Expt. c:} This experiment is the same as expt. b, but with two modifications. First, a penalty is added each time a camera is added to the rig to incentivize the imaging designer to not place unnecessary cameras. Second, the distribution of cars during training is changed to only be in front of the ego-vehicle to demonstrate that the imaging designer can customize its design to specific scenarios.

\begin{itemize}
\item \textbf{Expt. a:} The CD exercises a limited action space, including only $(p,z,\beta,\lambda)$ for a single camera on the front of the ego-vehicle. We use the same formulation as described above, but, at each step, if the CD places a camera, the previous camera is \emph{replaced}, rather than the new one being added on the rig. After six steps, the episode terminates.
%where to place a single camera in the front of the ego-vehicle, selecting only the pitch, height, and FoV. 
\item \textbf{Expt. b:} The CD exercises the full action space, including $(p,x,y,z,\theta,\beta,\lambda)$. Cameras are placed within a bounding box on top of the ego-vehicle, as shown in Fig.~\ref{fig:carla_overview}. For comparison with the nuScenes rig, which has six cameras, we set the episode length to six, so at most six cameras can be placed. 
\item \textbf{Expt. c:} This experiment includes two modifications to Expt. b. First, a penalty is enforced each time a camera is added to the rig to disincentivize the CD from placing unnecessary cameras. Second, the distribution of vehicles during training is changed to only be in front of the ego-vehicle to demonstrate that the CD can customize its rig design to specific scenarios.
\end{itemize}

We collect data on a Tesla Model 3 (TM3) since Renault Zoe (RZ) is not available in CARLA (placing cameras within the bounds of the RZ roof). Since TM3 is slightly smaller than RZ, this does not significantly affect what the cameras see. Our approach is flexible and the action space can be changed or other constraints added per requirements.

%one with a simple one camera rig where the imaging designer only optimizes the camera extrinsics and intrinsics and one where the imaging designer may choose how many cameras are placed. For both experiments, we set the episode to terminate after six steps. At each step of 

%The goal of our experiments is to show that the CD can find optimal camera rigs for the task of BEV segmentation. 

\vspace{1mm}
\noindent
\textbf{Evaluation Protocol:} After training, we use the following protocol to evaluate the quality of the CD-optimized camera rigs. First, we test the CD over 100 episodes, saving the candidate camera rig and sum of rewards at the end of each episode. We then select the top 20 rigs based on their sum of rewards. We fix these rigs and evaluate them over more episodes (20), again recording their sum of rewards. We sort the top twenty rigs by their sum of rewards and select the rig with the top reward, which we call the selected rig. % which we call the finalist.

We test the efficacy of the selected rig by comparing its BEV performance to that of the nuScenes \cite{caesar2020nuscenes} rig. To compare BEV performance, we collect 25,000 training images and 5,000 test images in CARLA using both our selected rig and the nuScenes rig. We do this by deploying both rigs on a Tesla Model 3. Next, we train one BEV segmentation model for each rig, using the collected training data. Finally, we test both BEV segmentation models on the corresponding test dataset captured from that rig. By collecting train and test data in the exact same CARLA scenes, we ensure a fair comparison. The test IoU then serves as a final measure of the selected rig's utility for BEV segmentation.
% In CARLA, we then deploy a vehicle (Audi TT, which has a similar size to the nuScenes vehicle) with both the top rig from the imaging designer and the standard nuScenes rig. The vehicle collects 25,000 training images and 5,000 testing images from each rig. 

\subsubsection{Results and Discussion}

%\vspace{1mm}
%\noindent
As shown in Fig.~\ref{fig:carla_results}, the CD significantly outperforms random search, and we observe that the rewards consistently increase over time across experiments. While using the pre-trained, frozen PM allows the CD to create camera configurations that increase BEV segmentation accuracy, jointly training the PM and CD together yields the best results. We note that the pre-trained CVT model (before RL) has 9\% and 11\% IoU for expts a and b, respectively, due to the challenging nature of fitting across many rigs. This IoU is improved during joint training. Using the above evaluation protocol for our CD, we find rigs created with the CD, for both experiments a and b, outperform the nuScenes rig on the task of BEV segmentation in our CARLA environment, as shown in Table \ref{table:iou}.  The top rigs for each experiment, example images from our rig vs. nuScenes, and a PM prediction are shown in Fig.~\ref{fig:carla_overview}. In expt. b, the CD can place up to the number of cameras in nuScenes (six). We find that the created rigs conform with AV conventions in several ways, such as distributing views around the ego-vehicle and using varying FoVs. While AV rigs, such as nuScenes, are well-engineered, our method suggests it may be possible to further improve them for specific tasks and environments.

%The main differences between our rigs and the nuScenes rig are camera placement (increasing height and decreasing pitch) and increasing overlap between views.

In expt. b, the CD learns to place the maximum number of cameras (six) on the rig since there is no penalty for placing additional cameras. However, in many cases, AV companies may want to reduce rig cost and inference time by using fewer cameras. Different camera rigs may also be better suited to different AV scenarios. We test whether the CD can take both of these considerations into account in expt. c by only placing cars in front of the ego-vehicle and enforcing a penalty to the reward each time an additional camera is added to the rig. As a result, we find that the CD places fewer cameras and places them facing forward, as shown in Fig.~\ref{fig:carla_results}. This result demonstrates that our approach can be used to build resource limited imaging systems that are well suited for specific test scenarios.

%\subsubsection{Discussion}

\vspace{1mm}
\noindent
\textbf{Strategy \#1 - Camera Placement:} Across experiments, we observe that the CD consistently learned two behaviors that lead to increased performance: (1) maximize camera height to 0.5 m above the ego-vehicle, and (2) reduce camera pitch to $-20^{\circ}$. Maximizing camera height reduces the number of occlusions, thus leading to more ground truth pixels, and potentially incentivizing this behavior. However, we note the average number of 3D bounding box labels across both test sets is the same, suggesting occlusions do not incentivize higher camera placement and BEV segmentation performance is naturally improved with higher camera positions. The negative pitch could mean the CD has learned to prioritize detecting nearby cars, perhaps because the perception model has higher confidence of those predictions. We also observe that all vehicles in the scene are still visible with a $-20^{\circ}$ pitch, and only the sky is cropped, thus the CD reduces the number of uninformative pixels, while maximizing the number of pixels on the road. Finally, we find that two front-facing cameras are placed at the rear of the vehicle in expt. b. We ablate this placement by re-training the PM with both cameras moved to the front. IoU is 5\% better when the cameras are in the back, perhaps since, by placing the front-facing cameras at the back, both the front and sides of the ego-vehicle are visible in captured images.

%the CD may have also learned to reduce the number of unimfortative pixels, while maximizing the number of pixels on the road.
%which could incentivize the CD to increase camera height. However, we note the average number of 3D bounding box labels across both test sets is the same, indicating occlusions are not responsible for the  overall BEV segmentation performance. 

\vspace{1mm}
\noindent
\textbf{Strategy \#2 - FoV vs Object Resolution:} In expt. a, we found the FoV was always maximized by the CD, which makes sense because it allows the CD to obtain higher reward when more vehicles in the scene are visible. In expt. b, the FoV of the target rig varies between $85^{\circ}$ and $120^{\circ}$, suggesting when all of the scene is visible (as is the case in expt. b due to the CD learning to distribute the camera yaws in all directions), FoV is less important or that the CD may have learned a tradeoff between FoV and object resolution.

%the cameras have full coverage of the scene (as is the case in expt. b due to the CD placing cameras facing all directions), FoV is less important or that the CD may have learned a tradeoff between FoV and object resolution.  

%In expt. c, we find that the imaging designer has learned both to place fewer cameras due to the penalty added each time a camera is added and learned to focus the cameras on the part of the scene where cars are placed. In this experiment, we only place cars in front of the ego-vehicle and, as a result, the imaging designer learns to place three.

%\vspace{1mm}
%\noindent
%\textbf{Real-World Deployment:} Our experiments are demonstrated in simulation, but have a direct path to real-world use. Our method can be used by AV companies to design rigs using their own simulator and internal requirements; those rigs can then be deployed on test cars to collect data. As AV simulators continue to improve, we expect any gap in rig utility in sim vs real to fall.

\vspace{1mm}
\noindent
\textbf{Limitations:} We demonstrate the CD on a limited number of scenarios within CARLA and focus only on the task of BEV segmentation of vehicles. In the future, our approach can be applied to more scenarios, tasks, and object classes. In addition, our experiments are done in simulation only. That said, our method has a direct path to real-world use. It can be used by AV companies to design rigs using their own simulator and requirements; those rigs can then be deployed on test cars to collect data. As AV simulators improve, we expect any gap in rig utility in sim vs real to fall.

% The objects in the environment are limited to vehicles but can be easily extended to include other classes as well. 

%-------------------------------------------------------------------------
%\section{Discussion}

%-------------------------------------------------------------------------
\section{Conclusion}

Our paper proposes a novel method to co-design camera configurations with perception models (PMs) for task-specific applications. We define a context-free grammar (CFG) that serves as a search space and theoretical framework for imaging system design. We then propose a camera designer (CD) that uses reinforcement learning to co-learn a camera configuration and PM for the proposed task by transforming the CFG into a state-action space. The PM is jointly trained with the CD and predicts the task output, which is used to compute the PM loss and reward for the CD to propose better candidate camera configurations. We demonstrate our method for co-design by applying it to (1) depth estimation using stereo cues, and (2) optimizing camera rigs for autonomous vehicle perception. We show in both cases that CD and PM can be learned together. Our co-design framework shows that camera configurations and perception models are closely linked and task-specific optimal designs that outperform human designs can be searched for computationally. 
% optimal designs that outperform human designed ones exist and can be search for computationally.

% We demonstrate that our approach can learn the fundamental vision technique of stereo depth estimation from scratch by jointly creating an imaging system and perception model. Moreover, our co-design framework can be used for practical engineering problems, such as optimizing camera rigs for AV perception tasks. We demonstrate that the perception model and imaging system (camera rig) can be jointly optimized for the task of BEV segmentation, leading to better downstream performance than achieved with rigs used in industry.

%a fundamental vision technique from scratch such as an imaging system and a perception model that estimates depth using stereo cues. Moreover, our co-design framework can also learn optimal imaging setups and perception model that outperform industry standards for BEV segmentation in autonomous vehicles. Our co-design framework shows that the imaging setup and a learned perception model are very closely linked and optimal designs that outperform human designed ones exist and can be search for computationally. 

% optimized camera rigs for autonomous vehicles, which lead to higher BEV segmentation performance than rigs currently used in industry in our simulated environment.

\vspace{2mm}
\noindent
\textbf{Acknowledgements:} We would like to thank Siddharth Somasundaram for his diligent proofreading of the paper. KT was supported by the SMART Contract IARPA Grant \#2021-20111000004. We would also like to thank Systems \& Technology Research (STR).

%-------------------------------------------------------------------------

{\small
\bibliographystyle{ieee_fullname}
\bibliography{egbib}

\begin{thebibliography}{10}\itemsep=-1pt

\bibitem{Arnab2021ViViTAV}
Anurag Arnab, Mostafa Dehghani, Georg Heigold, Chen Sun, Mario Lucic, and Cordelia Schmid.
\newblock Vivit: A video vision transformer.
\newblock {\em 2021 IEEE/CVF International Conference on Computer Vision (ICCV)}, pages 6816--6826, 2021.

\bibitem{baek2021polka}
Seung-Hwan Baek and Felix Heide.
\newblock Polka lines: Learning structured illumination and reconstruction for active stereo.
\newblock In {\em Proceedings of the IEEE/CVF Conference on Computer Vision and Pattern Recognition}, pages 5757--5767, 2021.

\bibitem{bakerdesigning}
Bowen Baker, Otkrist Gupta, Nikhil Naik, and Ramesh Raskar.
\newblock Designing neural network architectures using reinforcement learning.
\newblock In {\em International Conference on Learning Representations}.

\bibitem{bergstra2012random}
James Bergstra and Yoshua Bengio.
\newblock Random search for hyper-parameter optimization.
\newblock {\em Journal of machine learning research}, 13(2), 2012.

\bibitem{bhandari2022computational}
Ayush Bhandari, Achuta Kadambi, and Ramesh Raskar.
\newblock {\em Computational Imaging}.
\newblock MIT Press, 2022.

\bibitem{bradski2000opencv}
Gary Bradski.
\newblock The opencv library.
\newblock {\em Dr. Dobb's Journal: Software Tools for the Professional Programmer}, 25(11):120--123, 2000.

\bibitem{browne2012survey}
Cameron~B Browne, Edward Powley, Daniel Whitehouse, Simon~M Lucas, Peter~I Cowling, Philipp Rohlfshagen, Stephen Tavener, Diego Perez, Spyridon Samothrakis, and Simon Colton.
\newblock A survey of monte carlo tree search methods.
\newblock {\em IEEE Transactions on Computational Intelligence and AI in games}, 4(1):1--43, 2012.

\bibitem{caesar2020nuscenes}
Holger Caesar, Varun Bankiti, Alex~H Lang, Sourabh Vora, Venice~Erin Liong, Qiang Xu, Anush Krishnan, Yu Pan, Giancarlo Baldan, and Oscar Beijbom.
\newblock nuscenes: A multimodal dataset for autonomous driving.
\newblock In {\em Proceedings of the IEEE/CVF conference on computer vision and pattern recognition}, pages 11621--11631, 2020.

\bibitem{chakrabarti2016learning}
Ayan Chakrabarti.
\newblock Learning sensor multiplexing design through back-propagation.
\newblock {\em Advances in Neural Information Processing Systems}, 29, 2016.

\bibitem{chang2018hybrid}
Julie Chang, Vincent Sitzmann, Xiong Dun, Wolfgang Heidrich, and Gordon Wetzstein.
\newblock Hybrid optical-electronic convolutional neural networks with optimized diffractive optics for image classification.
\newblock {\em Scientific reports}, 8(1):1--10, 2018.

\bibitem{chang2019deep}
Julie Chang and Gordon Wetzstein.
\newblock Deep optics for monocular depth estimation and 3d object detection.
\newblock In {\em Proceedings of the IEEE/CVF International Conference on Computer Vision}, pages 10193--10202, 2019.

\bibitem{del2020learned}
Philipp Del~Hougne, Mohammadreza~F Imani, Aaron~V Diebold, Roarke Horstmeyer, and David~R Smith.
\newblock Learned integrated sensing pipeline: Reconfigurable metasurface transceivers as trainable physical layer in an artificial neural network.
\newblock {\em Advanced Science}, 7(3):1901913, 2020.

\bibitem{dosovitskiy2020image}
Alexey Dosovitskiy, Lucas Beyer, Alexander Kolesnikov, Dirk Weissenborn, Xiaohua Zhai, Thomas Unterthiner, Mostafa Dehghani, Matthias Minderer, Georg Heigold, Sylvain Gelly, et~al.
\newblock An image is worth 16x16 words: Transformers for image recognition at scale.
\newblock {\em arXiv preprint arXiv:2010.11929}, 2020.

\bibitem{dosovitskiy2017carla}
Alexey Dosovitskiy, German Ros, Felipe Codevilla, Antonio Lopez, and Vladlen Koltun.
\newblock Carla: An open urban driving simulator.
\newblock In {\em Conference on robot learning}, pages 1--16. PMLR, 2017.

\bibitem{cfg1}
Iddo Drori, Yamuna Krishnamurthy, Raoni Louren{\c{c}}o, R{\'{e}}mi Rampin, Kyunghyun Cho, Cl{\'{a}}udio~T. Silva, and Juliana Freire.
\newblock Automatic machine learning by pipeline synthesis using model-based reinforcement learning and a grammar.
\newblock {\em CoRR}, abs/1905.10345, 2019.

\bibitem{unrealengine}
{Epic Games}.
\newblock Unreal engine.

\bibitem{fawzi2022discovering}
Alhussein Fawzi, Matej Balog, Aja Huang, Thomas Hubert, Bernardino Romera-Paredes, Mohammadamin Barekatain, Alexander Novikov, Francisco~J R~Ruiz, Julian Schrittwieser, Grzegorz Swirszcz, et~al.
\newblock Discovering faster matrix multiplication algorithms with reinforcement learning.
\newblock {\em Nature}, 610(7930):47--53, 2022.

\bibitem{geary2002introduction}
Joseph~M Geary.
\newblock {\em Introduction to lens design: with practical ZEMAX examples}.
\newblock Willmann-Bell Richmond, VA, USA:, 2002.

\bibitem{guo2022polygrammar}
Minghao Guo, Wan Shou, Liane Makatura, Timothy Erps, Michael Foshey, and Wojciech Matusik.
\newblock Polygrammar: grammar for digital polymer representation and generation.
\newblock {\em Advanced Science}, 9(23):2101864, 2022.

\bibitem{guo2022data}
Minghao Guo, Veronika Thost, Beichen Li, Payel Das, Jie Chen, and Wojciech Matusik.
\newblock Data-efficient graph grammar learning for molecular generation.
\newblock {\em arXiv preprint arXiv:2203.08031}, 2022.

\bibitem{haas2014history}
John~K Haas.
\newblock A history of the unity game engine.
\newblock 2014.

\bibitem{haim2018depth}
Harel Haim, Shay Elmalem, Raja Giryes, Alex~M Bronstein, and Emanuel Marom.
\newblock Depth estimation from a single image using deep learned phase coded mask.
\newblock {\em IEEE Transactions on Computational Imaging}, 4(3):298--310, 2018.

\bibitem{he2018learning}
Lei He, Guanghui Wang, and Zhanyi Hu.
\newblock Learning depth from single images with deep neural network embedding focal length.
\newblock {\em IEEE Transactions on Image Processing}, 27(9):4676--4689, 2018.

\bibitem{jumper2021highly}
John Jumper, Richard Evans, Alexander Pritzel, Tim Green, Michael Figurnov, Olaf Ronneberger, Kathryn Tunyasuvunakool, Russ Bates, Augustin {\v{Z}}{\'\i}dek, Anna Potapenko, et~al.
\newblock Highly accurate protein structure prediction with alphafold.
\newblock {\em Nature}, 596(7873):583--589, 2021.

\bibitem{cfg3}
Michael Katz, Parikshit Ram, Shirin Sohrabi, and Octavian Udrea.
\newblock Exploring context-free languages via planning: The case for automating machine learning.
\newblock {\em Proceedings of the International Conference on Automated Planning and Scheduling}, 30(1):403--411, Jun. 2020.

\bibitem{klinghoffer2022physics}
Tzofi Klinghoffer, Siddharth Somasundaram, Kushagra Tiwary, and Ramesh Raskar.
\newblock Physics vs. learned priors: Rethinking camera and algorithm design for task-specific imaging.
\newblock In {\em 2022 IEEE International Conference on Computational Photography (ICCP)}, pages 1--12. IEEE, 2022.

\bibitem{uct}
Levente Kocsis and Csaba Szepesv{\'a}ri.
\newblock Bandit based monte-carlo planning.
\newblock In Johannes F{\"u}rnkranz, Tobias Scheffer, and Myra Spiliopoulou, editors, {\em Machine Learning: ECML 2006}, pages 282--293, Berlin, Heidelberg, 2006. Springer Berlin Heidelberg.

\bibitem{land2012animal}
Michael~F Land and Dan-Eric Nilsson.
\newblock {\em Animal eyes}.
\newblock OUP Oxford, 2012.

\bibitem{levine2020offline}
Sergey Levine, Aviral Kumar, George Tucker, and Justin Fu.
\newblock Offline reinforcement learning: Tutorial, review, and perspectives on open problems.
\newblock {\em arXiv preprint arXiv:2005.01643}, 2020.

\bibitem{pyredner}
Tzu-Mao Li, Miika Aittala, Fr{\'e}do Durand, and Jaakko Lehtinen.
\newblock Differentiable monte carlo ray tracing through edge sampling.
\newblock {\em ACM Trans. Graph. (Proc. SIGGRAPH Asia)}, 37(6):222:1--222:11, 2018.

\bibitem{luo2016efficient}
Wenjie Luo, Alexander~G Schwing, and Raquel Urtasun.
\newblock Efficient deep learning for stereo matching.
\newblock In {\em Proceedings of the IEEE conference on computer vision and pattern recognition}, pages 5695--5703, 2016.

\bibitem{cfg2}
Radu Marinescu, Akihiro Kishimoto, Parikshit Ram, Ambrish Rawat, Martin Wistuba, Paulito~P. Palmes, and Adi Botea.
\newblock Searching for machine learning pipelines using a context-free grammar.
\newblock {\em Proceedings of the AAAI Conference on Artificial Intelligence}, 35(10):8902--8911, May 2021.

\bibitem{mazyavkina2021reinforcement}
Nina Mazyavkina, Sergey Sviridov, Sergei Ivanov, and Evgeny Burnaev.
\newblock Reinforcement learning for combinatorial optimization: A survey.
\newblock {\em Computers \& Operations Research}, 134:105400, 2021.

\bibitem{metzler2020deep}
Christopher~A Metzler, Hayato Ikoma, Yifan Peng, and Gordon Wetzstein.
\newblock Deep optics for single-shot high-dynamic-range imaging.
\newblock In {\em Proceedings of the IEEE/CVF Conference on Computer Vision and Pattern Recognition}, pages 1375--1385, 2020.

\bibitem{mirhoseini2021graph}
Azalia Mirhoseini, Anna Goldie, Mustafa Yazgan, Joe~Wenjie Jiang, Ebrahim Songhori, Shen Wang, Young-Joon Lee, Eric Johnson, Omkar Pathak, Azade Nazi, et~al.
\newblock A graph placement methodology for fast chip design.
\newblock {\em Nature}, 594(7862):207--212, 2021.

\bibitem{onzon2021neural}
Emmanuel Onzon, Fahim Mannan, and Felix Heide.
\newblock Neural auto-exposure for high-dynamic range object detection.
\newblock In {\em Proceedings of the IEEE/CVF Conference on Computer Vision and Pattern Recognition}, pages 7710--7720, 2021.

\bibitem{robidoux2021end}
Nicolas Robidoux, Luis E~Garcia Capel, Dong-eun Seo, Avinash Sharma, Federico Ariza, and Felix Heide.
\newblock End-to-end high dynamic range camera pipeline optimization.
\newblock In {\em Proceedings of the IEEE/CVF Conference on Computer Vision and Pattern Recognition}, pages 6297--6307, 2021.

\bibitem{schaeffer1996new}
Jonathan Schaeffer and Aske Plaat.
\newblock New advances in alpha-beta searching.
\newblock In {\em Proceedings of the 1996 ACM 24th annual conference on Computer science}, pages 124--130, 1996.

\bibitem{schulman2017proximal}
John Schulman, Filip Wolski, Prafulla Dhariwal, Alec Radford, and Oleg Klimov.
\newblock Proximal policy optimization algorithms.
\newblock {\em arXiv preprint arXiv:1707.06347}, 2017.

\bibitem{selva2022video}
Javier Selva, Anders~S Johansen, Sergio Escalera, Kamal Nasrollahi, Thomas~B Moeslund, and Albert Clap{\'e}s.
\newblock Video transformers: A survey.
\newblock {\em arXiv preprint arXiv:2201.05991}, 2022.

\bibitem{sitzmann2018end}
V. Sitzmann, S. Diamond, Y. Peng, X. Dun, S. Boyd, W. Heidrich, F. Heide, and G. Wetzstein.
\newblock End-to-end optimization of optics and image processing for achromatic extended depth of field and super-resolution imaging.
\newblock {\em ACM Trans. Graph. (SIGGRAPH)}, 2018.

\bibitem{sun2020learning}
Qilin Sun, Ethan Tseng, Qiang Fu, Wolfgang Heidrich, and Felix Heide.
\newblock Learning rank-1 diffractive optics for single-shot high dynamic range imaging.
\newblock In {\em Proceedings of the IEEE/CVF conference on computer vision and pattern recognition}, pages 1386--1396, 2020.

\bibitem{tseng2021differentiable}
Ethan Tseng, Ali Mosleh, Fahim Mannan, Karl St-Arnaud, Avinash Sharma, Yifan Peng, Alexander Braun, Derek Nowrouzezahrai, Jean-Francois Lalonde, and Felix Heide.
\newblock Differentiable compound optics and processing pipeline optimization for end-to-end camera design.
\newblock {\em ACM Transactions on Graphics (TOG)}, 40(2):1--19, 2021.

\bibitem{vazquez2022gramml}
Hernan~Ceferino Vazquez, Jorge S{\'a}nchez, and Rafael Carrascosa.
\newblock Gram{ML}: Exploring context-free grammars with model-free reinforcement learning.
\newblock In {\em Sixth Workshop on Meta-Learning at the Conference on Neural Information Processing Systems}, 2022.

\bibitem{zhang2022deep}
Tianyu Zhang, Amin Banitalebi-Dehkordi, and Yong Zhang.
\newblock Deep reinforcement learning for exact combinatorial optimization: Learning to branch.
\newblock In {\em 2022 26th International Conference on Pattern Recognition (ICPR)}, pages 3105--3111. IEEE, 2022.

\bibitem{zhao2022automatic}
Allan Zhao, Tao Du, Jie Xu, Josie Hughes, Juan Salazar, Pingchuan Ma, Wei Wang, Daniela Rus, and Wojciech Matusik.
\newblock Automatic co-design of aerial robots using a graph grammar.
\newblock In {\em 2022 IEEE/RSJ International Conference on Intelligent Robots and Systems (IROS)}, pages 11260--11267. IEEE, 2022.

\bibitem{zhou2022cross}
Brady Zhou and Philipp Kr{\"a}henb{\"u}hl.
\newblock Cross-view transformers for real-time map-view semantic segmentation.
\newblock In {\em Proceedings of the IEEE/CVF Conference on Computer Vision and Pattern Recognition}, pages 13760--13769, 2022.

\bibitem{zoph2016neural}
Barret Zoph and Quoc~V Le.
\newblock Neural architecture search with reinforcement learning.
\newblock {\em arXiv preprint arXiv:1611.01578}, 2016.

\end{thebibliography}
}

\end{document}